\documentclass[10pt,twocolumn,letterpaper]{article}

\usepackage{cvpr}              
\usepackage{placeins}
\usepackage{footmisc} 
\usepackage{multirow}
\usepackage{booktabs}
\usepackage{multirow}
\usepackage{graphicx} 

\usepackage[dvipsnames]{xcolor}

\definecolor{cvprblue}{rgb}{0.21,0.49,0.74}
\usepackage[pagebackref,breaklinks,colorlinks,citecolor=cvprblue]{hyperref}

\def\paperID{90} 
\def\confName{3DV\xspace}
\def\confYear{2025\xspace}

\title{HoloGest: Decoupled Diffusion and Motion Priors for Generating Holisticly Expressive Co-speech Gestures}

\author{Yongkang Cheng\\
Tencent AILab\\
Shenzhen, Guangdong\\
{\tt\small cyk19990422@gmail.com}
\and
Shaoli Huang\footnotemark[1]\\ 
Tencent AILab\\
Shenzhen, Guangdong\\
{\tt\small shaol.huang@gmail.com}\\
}

\begin{document}

\twocolumn[{
\maketitle
\begin{center}
    \captionsetup{type=figure}
    \includegraphics[width=1.\textwidth]{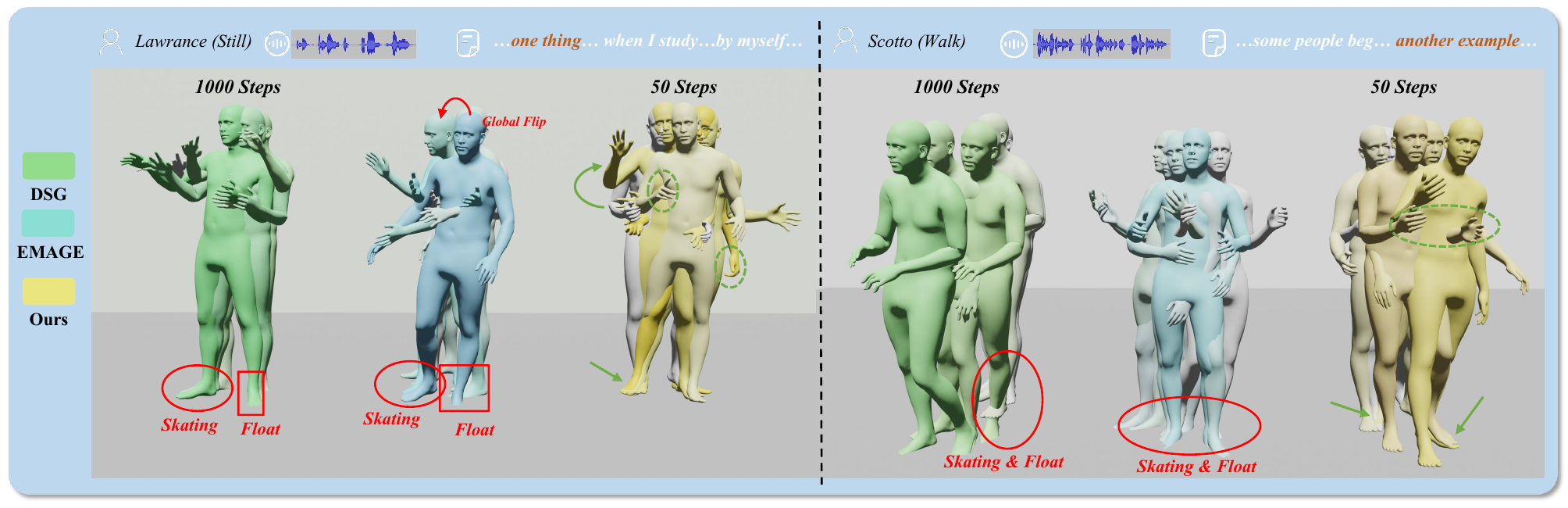}
    \captionof{figure}{ 
A comparison of three methods: DSG, a diffusion-based co-speech gesture generation method using DDPM (stiff limbs, slow inference, physically unnatural); EMAGE, an autoregressive generation method using VAE (motion artifacts, global flipping, physically unnatural); and our proposed generation method (rich movements, lively fingers, physically natural). The transition from past frames to the current frame (every 10 frames) is represented by the gradient in virtual human color, from light to dark.} 
 \label{fig:teaser}
\end{center}
}]
\footnotetext[1]{$*$ means Corresponding author.}

\begin{abstract}
Animating virtual characters with holistic co-speech gestures is a challenging but critical task. Previous systems have primarily focused on the weak correlation between audio and gestures, leading to physically unnatural outcomes that degrade the user experience. To address this problem, we introduce \textbf{HoleGest}, a novel neural network framework based on decoupled diffusion and motion priors for the automatic generation of high-quality, expressive co-speech gestures. Our system leverages large-scale human motion datasets to learn a robust prior with low audio dependency and high motion reliance, enabling stable global motion and detailed finger movements. To improve the generation efficiency of diffusion-based models, we integrate implicit joint constraints with explicit geometric and conditional constraints, capturing complex motion distributions between large strides. This integration significantly enhances generation speed while maintaining high-quality motion. Furthermore, we design a shared embedding space for gesture-transcription text alignment, enabling the generation of semantically correct gesture actions. Extensive experiments and user feedback demonstrate the effectiveness and potential applications of our model, with our method achieving a level of realism close to the ground truth, providing an immersive user experience. Our code, model, and demo are are available at https://cyk990422.github.io/HoloGest.github.io/.
\end{abstract}

\section{Introduction}
\label{sec:intro}

Animating virtual characters with holistic co-speech gestures is a challenging but critical task in various fields such as entertainment, education, and telecommunication. These gestures play a vital role in enhancing the naturalness and appeal of virtual characters, as they convey non-verbal information and improve the overall communication experience. However, generating holistic co-speech gestures that accurately represent the complex interplay between audio and body motion remains a challenging task, primarily due to the weak correlation between audio and both global motion trajectory and finger movements. This weak correlation often leads to physically unnatural outcomes, such as jittering in the global motion trajectory and poor expressiveness of finger motions, significantly reducing the effect of virtual characters.

Previous co-speech gesture generation methods can be divided into two categories: VAE-based or VQ-VAE-based generation systems~\cite{liu2022beat,liu2022audio} and diffusion-based generation systems~\cite{yang2024freetalker,zhu2023taming}. The former maps weakly correlated gesture-audio pairs to a low-dimensional latent space and learns a continuous probability distribution, from which new latent vectors are sampled and decoded to obtain co-speech gestures. However, due to the VAE's reconstruction loss~\cite{kingma2013auto} based on joint-level errors and the ambiguity in the latent space, the generated gestures often appear overly smooth or unnatural. In contrast, diffusion-based methods model gesture generation as a gradual diffusion process, where the mapping relationship between audio and gestures is gradually established through a series of noise diffusion steps. Compared to VAE, this method can generate gesture sequences with rich details while maintaining audio synchrony. However, the high computational density and the resulting time cost limit the further development of diffusion methods. Furthermore, both approaches lack consideration of prior knowledge in motion, focusing only on the weakly correlated mapping between audio and upper-body gestures, and neglecting the physical laws of overall movement, such as continuity, stability, and rationality. Therefore, the generated holistic gestures may exhibit unnatural sliding, hovering phenomena, and monotonous finger movement, leading to a lack of overall naturalness and expressiveness, as shown in Figure~\ref{fig:teaser}.

To address these challenges, we introduce HoloGest, a novel diffusion-based framework for automatically synthesizing high-fidelity holistic co-speech gesture sequences from audio. Our system posits that in holistic co-speech gestures, limb movements are correlated with audio, global trajectories are related to limbs and independent of audio, and fingers are associated with both arm movements and audio information. Based on this, we learn a global trajectory diffusion generative prior model guided by limb movements on a large-scale human motion dataset. Simultaneously, we learn a finger diffusion generative prior model guided by the arms on a mixed sign language and gesture dataset, leaving audio and semantic features blank for subsequent fine-tuning. The former provides our system with strong locomotion prior, overcoming long-standing issues of unnatural sliding and jittering, while the latter offers more diverse finger movements, assisting in generating more vivid and high-fidelity gesture results.

Unlike previous methods that model the denoising process of the whole body as a single distribution, our system decouples the upper limbs, lower body, and fingers into three smaller and simpler subproblems, breaking down holistic co-speech gestures. During the denoising process, each sub-model focuses more on the distribution of specific body parts, thereby improving the generation quality of each part. However, the parallel diffusion denoising processes for the three parts further reduce the generation efficiency. To break free from this constraint, we employ a semi-implicit constraint~\cite{xu2023semi}, modeling large-stride complex multimodal distributions between adjacent denoising steps to significantly reduce the required number of denoising steps, thus achieving acceleration. 

Predicting gestures from speech is a challenging multimodal mapping task. A single speech segment can correspond to multiple gestures, making the association between the semantics intended to be conveyed in speech and gestures non-intuitive. Our system adopts the JEPA strategy~\cite{garrido2024learning,bardes2023v} to learn a gesture-speech joint embedding space. We first introduce wav2vec2~\cite{baevski2020wav2vec} for text transcription, then extract textual features and map them with gestures to a shared low-dimensional space based on a variational autoencoder. Finally, we introduce a predictor layer to further extract semantic features, aligning these abstract semantic features with the low-dimensional latent variables of gestures in this space. This approach maintains semantic alignment while generating natural gestures closely related to speech.

To demonstrate the inspirational value of motion priors in our system for the human motion generation domain, we further fine-tune our framework on the music-to-dance task, addressing the physical naturalness of generated results and showcasing its powerful generalization capabilities. To the best of our knowledge, our system represents the first audio-whole body gesture generation model considering motion priors, capable of generating high-fidelity, diverse, and physically natural holistic co-speech gesture sequences based on arbitrary user-provided audio (speech or music). We showcase our approach on multiple publicly available audio-motion datasets, and extensive experimental results indicate that, compared to VAE systems, our method generates more diverse and higher-quality results, while maintaining the naturalness of overall motion compared to diffusion systems, significantly reducing time costs and providing users with a novel experience. The importance of algorithmic design is also validated through ablation experiments.

\section{Related Work}
\label{sec:Related}

\textbf{Audio-to-motion Generation.} 
 Initial data-driven methods (~\cite{liu2022audio,habibie2021learning}) aimed to learn gesture matching from human demonstrations but lacked diversity. With the increasing interest in these action reconstruction methods~\cite{cheng2023bopr, liang2024ropetp, yu2024signavatars}, the training datasets have gradually become more abundant. Subsequent works (~\cite{habibie2021learning,yi2023generating,xie2022vector, cheng2024conditional, cheng2024expgest}) improved model diversity and introduced unique, expressive gestures. Some studies (~\cite{yang2023unifiedgesture,ahuja2020style,ao2023gesturediffuclip}) trained unified models for multiple speakers, embedding styles or applying style transfer techniques. Other research (~\cite{zhou2022gesturemaster,habibie2022motion}) utilized motion matching for gesture sequences, despite requiring complex rules. Audio-driven animation has gained attention, with virtual speaker animation advancements attributed to high-quality gesture datasets ZeroEGGs~\cite{ghorbani2023zeroeggs} and BEATX~\cite{liu2023emage}. Talkshow~\cite{yi2023generating} and EMAGE~\cite{liu2023emage} improved user experience by incorporating facial and expression parameters in virtual speaker generation. However, these methods face unnatural holistic gesture issues, and models lacking human motion knowledge struggle with physical problems like jittering, foot sliding, and floating, hindering the field's development. Some works are limited to the human upper body. HoloGest innovatively introduces a motion prior model to address these physical unnatural problems in gesture generation, providing a more engaging user experience.

\noindent\textbf{Diffusion Generative Models} have achieved remarkable results across various domains~\cite{rombach2022high,chen2023executing,saharia2022palette, han2024reindiffuse, han2024hutumotion}, especially in human motion generation. Motion Diffuse~\cite{zhang2022motiondiffuse} first applied diffusion models to text-conditioned human motion generation, offering fine-grained control of body parts. MDM~\cite{tevet2022human} is a milestone work using a motion diffusion model to manipulate motion representation based on input text control conditions. DSG~\cite{yang2023diffusestylegesture} generates well-matched results with speech using an attention mechanism. However, due to the high dimensionality and interactivity of diffusion models, motion generation based on the original diffusion model DDPM~\cite{ho2020denoising} suffers from time overhead. MLD~\cite{chen2023executing} introduces latent diffusion to motion generation, reducing computational resources and employing DDIM~\cite{song2020denoising} to enhance inference speed. Nevertheless, this two-stage method is non-end-to-end, and DDIM's noise step stacking and denoising step discarding result in artifacts. HoloGest addresses these issues by being the first method in gesture generation to use GAN~\cite{xie2021dual} for accelerating diffusion model inference speed. By increasing denoising step size and reducing denoising steps, it maintains high-quality diffusion model advantages while enabling rapid generation.

 \begin{figure*}
\center
  \includegraphics[width=\textwidth]{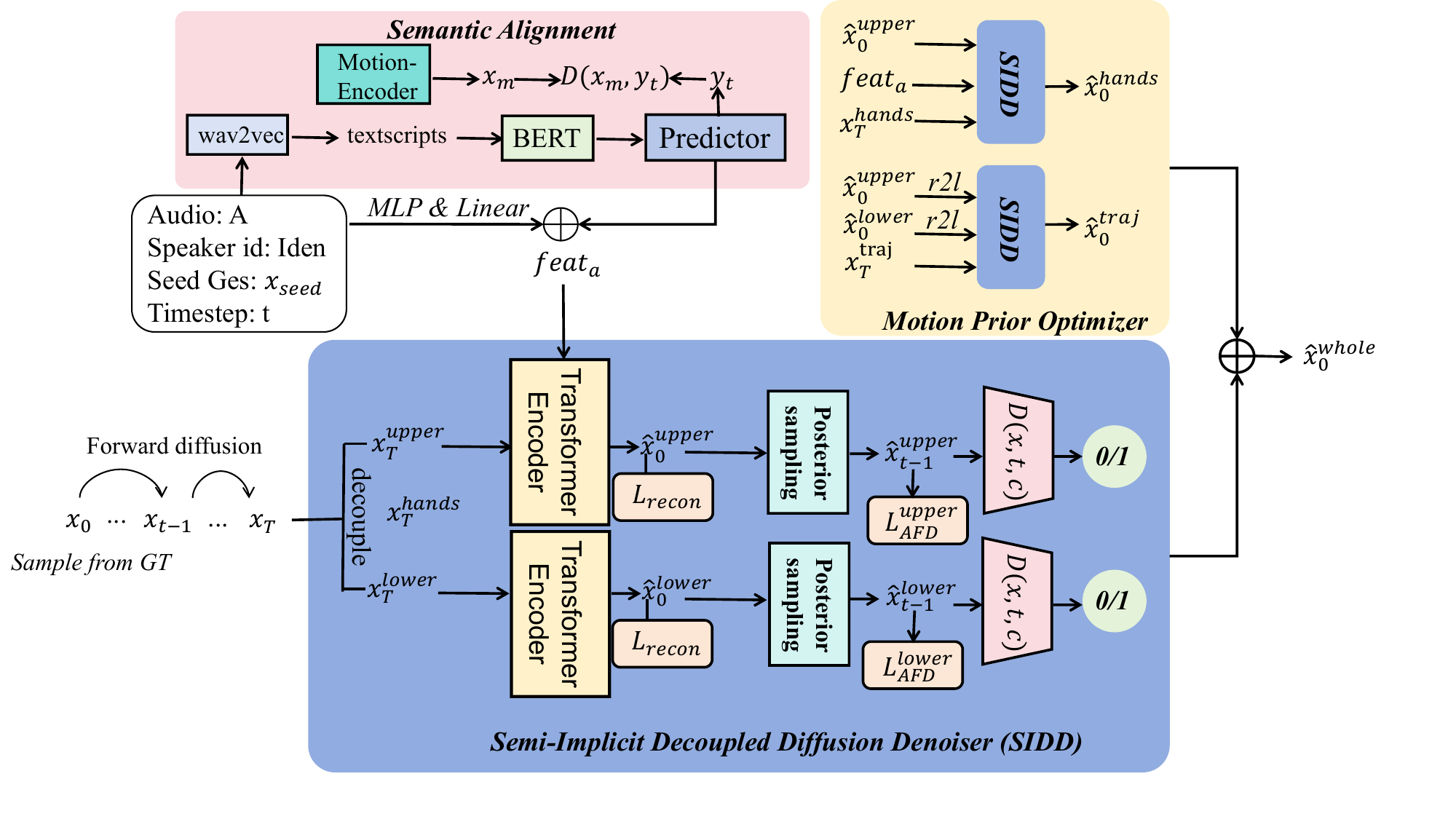}
  \caption{
  Our system comprises a semantic alignment module and two core components: (a) The semantic alignment module maps both the transcribed text and gesture sequence into the latent space simultaneously, further abstracting the semantic latent variables and aligning them with the gesture latent variables in a higher-level abstract space, serving as independent guiding tokens. (b) The semi-implicit decoupled denoiser, by introducing GAN and semi-implicit constraints, models the complex denoising distribution between adjacent large strides, accelerating generation by reducing the number of steps. (c) The motion prior optimization takes the denoised initial local gesture sequence as a condition, and in conjunction with the audio guiding signal, generates global motion and finger actions for the second time. This system requires no additional input and has no time constraints; any pure audio file can generate a set of vivid, natural, and high-quality holistic co-speech gesture sequences. 'r2l' represents converting the rotation representation to the coordinate representation using the SMPL model.
  }
  \label{fig:pip}
\end{figure*}

\section{Method}

\subsection{System Overview}
Our system synthesizes vivid, physically natural, and holistic co-speech gestures using only audio input. It is built on the human motion diffusion model (MDM) framework, employing the Diffusion Model to model adjacent denoising step distributions and supervising human geometric constraints for motion quality. The system structure, shown in Figure~\ref{fig:pip}, comprises two core components: (a) an end-to-end decoupled diffusion generative model that accepts audio input and denoises human joint sequences in parallel, and (b) a motion prior optimizer pre-trained on a large-scale human dataset, re-optimizing global motion and finger actions based on generated joint cues for natural and vivid virtual speakers. We also transcribe text and utilize the JEPA~\cite{assran2023self} strategy to extract semantic cues, enhancing result richness. To address DDPM denoising inefficiency, a semi-implicit denoising process is introduced for faster generation. In subsequent sections, we detail the system's key components.

\subsection{Decoupled Diffusion Denoiser}
\textbf{Brief overview of MDM.} Unlike traditional diffusion model-based methods, MDM~\cite{tevet2022human} considers the inherent physical constraints of the three-dimensional human body by predicting the original human motion representation instead of predicting noise, deviating from the DDPM process in conventional image generation. Therefore, at each step of the denoising process, MDM reconstructs the original representation from pure Gaussian noise, and ultimately generate the final result through the iterative process of noise addition and denoising:
\begin{equation}
    \hat{x}_{0}=\epsilon_{t}^{\theta}\left(x_{t}|c\right),\\
    x_{t-1}=\frac{1-\alpha_{t-1}+\sqrt{\alpha_{t-1}} \hat{x}_{0}}{1-\alpha_{t}\hat{x}_{0}}+\sigma_{t} z_{t},
\end{equation}
where $c$ is the control signal.

\noindent\textbf{Decoupled Denoiser Structure.} 
We construct our denoiser framework on MDM, tailoring it for the audio-to-gesture task with conditions including noise step , seed pose, audio information, and semantic latent code. The noise step  and seed pose are projected to the same dimension via MLP and linear layers, respectively, and subsequently added together. Audio is encoded using WavLM and time-dimension interpolated to align with gesture frames.

Although the denoising probability model generates satisfactory gestures, finger motion differs from limb motion. Limb movements exhibit larger amplitudes and correlate with melody, while finger movements are smaller, more precise, and semantically matched. Holistic modeling prioritizes body data matching over finger movements, reducing overall gesture expressiveness. 

To address this, we decouple the human body into upper limbs, lower limbs, and fingers, denoising these parts in parallel. However, the absence of global associations can result in unnatural motion, such as sudden orientation flips. \textit{To alleviate this, we concatenate the three-part features and map them to an independent conditional token, providing a global constraint for generating coherent results.}

\subsection{Semi-implicit Matching Constraint}
We've improved the network structure for better results, but DDPM's high computational complexity still limits diffusion generative methods' potential. This issue arises from DDPM's assumption that small, unimodal noise is added at each step, requiring many steps for denoising. Increasing noise step size disrupts the Gaussian distribution, making a simple $L2$ loss inadequate for modeling complex motion distribution and causing unnatural jittering.

To address this, we incorporate a GAN structure inspired by SiDDMs~\cite{xu2023semi} as an implicit objective to learn the denoiser. The GAN's conditional discriminator differentiates between the predicted denoising and original motion distributions, while the conditional denoiser aims to make them indistinguishable. The process is described by equation (2).
\begin{equation}
\begin{aligned}
    \underset{\theta}{\min} \underset{D_{adv}}{\max} \sum\limits_{t>0} \mathbb{E}_{q(x_{t})}D_{adv}(q(x_{t-1}|x_{t})||p_{\theta}(x_{t-1}|x_{t})),
\end{aligned}
\end{equation}
By examining the implementation of equation (2), it is clear that, during the adversarial stage, the method indirectly matches the conditional distribution by aligning with the joint distribution:
\begin{equation}
\begin{aligned}
    \underset{\theta}{\min} \underset{D_{adv}}{\max}  \mathbb{E}_{q(x_{0})q(x_{t-1}|x_{0})q(x_{t}|x_{t-1})}D_{adv}(q(x_{t-1},x_{t})||\\p_{\theta}(x_{t-1},x_{t})),
\end{aligned}
\end{equation}

However, adversarial training is a purely implicit matching process, typically used to constrain distributions that cannot be explicitly represented. We consider using a simpler marginal distribution to replace the joint distribution in equation (3). That is, we directly compute the posterior distribution, and then use the forward process for adversarial learning to model the large-step denoising distribution. The equation is represented as follows:
\begin{equation}
\begin{aligned}
    \underset{\theta}{\min} \underset{D_{\varnothing}}{\max}  
 ~\mathbb{E}_{q(x_{0})q(x_{t-1}|x_{0})q(x_{t}|x_{t-1})}[-log(D_{\varnothing}(x_{t-1},c,t))] \\
    +[-log(1-D_{\varnothing}(\hat{x}_{t-1},c,t))],
\end{aligned}
\end{equation}

Although we have simplified the implicit matching process, making adversarial training more stable, we have also encountered a new problem. Since the large-step denoising distribution is typically a complex multimodal distribution, the posterior sampling $p_{\theta}(\hat{x_{t-1}}|x_{t},\hat{x_{0}})$ result still has a significant difference from the forward sampling process, preventing our denoiser from successfully reversing from the pure noise distribution to the original distribution. Based on this, we employ the regularization term, Auxiliary Forward Diffusion Constraint (AFD), to explicitly constrain the similarity between the backward sampling results and forward diffusion results at the same time step. Its expression is as follows:
\begin{equation}
\begin{aligned}
    \mathbb{E}_{q(x_{0})q(x_{t-1}|x_{0})q(x_{t}|x_{t-1})}\frac{(1-\beta_{t})||\hat{x}_{t-1}-x_{t-1}||^{2}}{\beta_{t}},
\end{aligned}
\end{equation}
where $\sqrt{1-\beta_{t}}x_{t-1}$ represents the mean of the forward process $q(x_{t}|x_{t-1})$, and $\beta_{t}$ represents the variance table within the interval (0,1]. All models are trained using the AdamW optimizer with a fixed learning rate l. We apply EMA decay to the optimizer during the training process. The final training objective is:
\begin{equation}
\begin{aligned}
    \underset{\theta}{\min} \underset{D_{adv}}{\max}  \mathbb{E}_{q(x_{0})q(x_{t-1}|x_{0})q(x_{t}|x_{t-1})}[-log(D_{\varnothing}(x_{t-1},c,t))] \\
    +[-log(1-D_{\varnothing}(\hat{x}_{t-1},c,t))] + \lambda_{recon}\mathcal{L}_{Recon}\\
    + \lambda_{AFD}\frac{(1-\beta_{t})||\hat{x}_{t-1}-x_{t-1}||^{2}}{\beta_{t}},
\end{aligned}
\end{equation}
where $\lambda_{recon}$ represents the reconstruction weight of the denoiser, and $\lambda_{AFD}$ represents the weight of the regularization term. $\mathcal{L}_{Recon}$ represents the Mean Squared Error Loss between the denoised $\hat{x_{0}}$ and the original data $x_{0}$.

\subsection{Motion Prior Optimizer}
Despite the semi-implicit decoupled denoiser's ability to recover detailed and expressive gesture sequences, the correlation between audio and motion remains weak. This results in issues like foot sliding, jittering, and unnatural movements in previous methods. We believe the problem stems from the fact that not all aspects of co-speech gestures depend on audio. For instance, trajectories may not be related to audio beats but are closely connected to limb movements. To address these issues, we designed a motion prior optimizer.

\noindent\textbf{Trajectory Prior Optimizer.} 
In our system, the trajectory is more closely linked to limb posture than to audio. Therefore, we use limb posture as the guiding condition to recreate the global motion trajectory. Thanks to large-scale public human motion datasets, our model can acquire extensive motion prior knowledge. Taking a cue from GLAMR~\cite{yuan2022glamr}, we define the trajectory as a 9-dimensional parameter $G=(\Delta x, \Delta y, \Delta z, rot6d\in R^{6})$, where the last six dimensions represent global rotation and the first three dimensions represent displacement increments along the $XYZ$ axes, ensuring smoother results. 

The trajectory prior model continues to use the semi-implicit diffusion approach, with the 3D coordinates of the 21 human joint parts (excluding the root joint) and the time step $t$ as conditional guidance. These are independently mapped to conditional tokens and input into the Transformer-Encoder-based denoiser. The denoising process follows equation (6).

\noindent\textbf{Finger Prior Optimizer.} We've observed that finger movements guided solely by audio or semantic signals often lack dynamism and expressiveness. We believe finger movements correlate with forearm movements, such as a person pointing in a certain direction when raising their arm forward. Hence, we train finger priors on large-scale sign language and gesture datasets. During finger prior training, the guiding condition only uses the 6D rotation representation of the human forearm, leaving semantic and audio features empty for subsequent fine-tuning. The finger results are denoised using equation (3).

When fine-tuning on the BEATX dataset, we incorporate the audio signal into the finger prior's conditional guidance and link the entire system together for fine-tuning. During inference, the motion prior model serves as an optimizer, using the body generated by the semi-implicit decoupled denoiser as a condition, and generates the global trajectory and finger rotations as the final output.

\subsection{Semantic Alignment }
The existence of many-to-many mapping relationships between audio content and gesture sequences poses a significant challenge for generating semantically aligned actions accurately. To address this issue, we learn a joint embedding space for gestures and audio transcriptions, allowing them to align in an abstract space and reveal the semantic associations between the two modalities. Inspired by I-JEPA~\cite{assran2023self}, we initially train gesture and transcription encoders separately using a motion VAE structure and a BERT tokenizer~\cite{devlin2018bert}. We parameterize the transcriptions as tokenized word embedding sequences and linearly map them to a space with the same dimensions as the gesture latent codes $x_{m}$. Finally, we introduce a Predictor to further abstract semantic features $y_{t}$ from the latent space and fine-tune the encoders using CLIP-style contrastive learning $D(x_{m},y_{t})$. Both the motion VAE and Predictor structures adopt the traditional Transformer architecture. The NT-Xent~\cite{chen2020simple} loss is used in contrastive learning, with the goal of maximizing the similarity of transcription-gesture matched pairs in the latent space while minimizing the similarity of non-matched pairs. Formally, the loss function is as follows:
\begin{equation}
\begin{aligned}
    \mathcal{L}(t,m) = -log\frac{exp(sim(x_{t},y_{m})/\tau)}{\sum_{k\in K}exp(sim(x_{t},y_{m})/\tau)},\\
\end{aligned}
\end{equation}
where, $x_{t}$ and $y_{m}$ are the latent space representations of a matching transcription-gesture pair. $sim$ is similarity score between two latent codes, $K$ is a set containing one positive sample transcription and a group of negative sample gestures, and $\tau$ is the temperature parameter used to adjust the sensitivity of the function. Finally, we freeze the trained semantic alignment module and deploy only the transcription encoder into the system, ensuring that the final generated results accurately capture the semantic content.

\section{Experiments}
In this section, we evaluate the effectiveness of the proposed system in generating holistic co-speech gestures from audio and compare it with contemporary holistic gesture generation methods to demonstrate the superiority of our system. Ablation studies further validate the roles of essential modules and design choices within the system. Generalization experiments showcase the potential value and application prospects of our proposed method in this domain. Considering the subtle nature of human gestures for evaluation, we conduct extensive user studies to substantiate the superior performance of the proposed system. \textbf{\textit{\textcolor{black}{We strongly encourage readers to refer to the accompanying video for additional qualitative evaluations and application results.}}}
\subsection{Experiment Design}
\textbf{Datasets.} For the audio-independent global trajectory motion prior module, we train on the 100-STYLE~\cite{mason2018style} and AMASS~\cite{mahmood2019amass} datasets. Both are large-scale publicly available Mocap datasets, with the former containing over 4 million frames of 100 different locomotion styles, and the latter being a large-scale human motion dataset, both represented with 55 joints in SMPLX~\cite{loper2015smpl} format. The trajectory prior is trained using all datasets. For the finger prior module associated with arm movements, we train on the SignAvatars~\cite{yu2023signavatars} sign language dataset and the audio-removed BEATX~\cite{liu2023emage} dataset. The former contains SMPLX representations of multiple sign language videos shared with us by the authors, and the latter is a publicly available large-scale gesture dataset, uniformly represented in SMPLX format, containing 24 English speakers. When training the finger prior, we mix all speaker data and sign language data for training. Finally, we train the audio-to-gesture model on the BEATX dataset, and during the fine-tuning of fingers, we release the audio features and semantic alignment as additional guiding signals to generate natural and rich finger movements. We evaluate the model's effectiveness on the BEATX test set.

\noindent\textbf{Evaluation Metrics.} To evaluate the effectiveness of our proposed system, in addition to focusing on the common Frechet Gesture Distance (FGD)~\cite{yoon2022genea}, Beat Alignment (BA)~\cite{liu2023emage}, and Diversity (DIV)~\cite{liu2022beat} metrics, we also introduce physical naturalness evaluation metrics, including Skating (Skate)~\cite{karunratanakul2023guided} and Floating (Float)~\cite{yuan2023physdiff}, and define a Semantic Alignment score (SA) to validate the performance of the semantic alignment module. The first three are used to evaluate the quality of generated gestures: (1) \textit{FGD} is a common metric in generative models, used to evaluate the difference between the distribution of generated movements and the original training distribution, providing insights into the fidelity and similarity between generated data and real data. (2) \textit{BA} is used to evaluate the synchronicity of speech and movement, with higher values indicating better alignment with the audio beat. (3) \textit{DIV} measures the L1 distance between multiple body gestures generated under the same control signal, with larger values indicating greater diversity.

To evaluate the physicality of holistic co-speech gestures, we use (4) \textit{Skate} to quantify the displacement distance of the virtual character's toes when their feet are in contact with the ground (determined by setting a toe acceleration threshold). This is crucial for the naturalness and authenticity of overall motion, as realistic motion results can provide users with an immersive experience. (5) \textit{Float} is used to assess the floating distance of the virtual character along the y-axis. We assume the ground level to be the lowest point of the sequence plus 0.5 cm, and when the character has at least one foot in contact with the ground, we calculate the distance between the toes and the ground to quantify the degree of floating.

To evaluate the semantic consistency between speech and generated gestures, we define a new metric called Semantic Alignment (SA)~\cite{ao2023gesturediffuclip}. It assesses the degree of semantic alignment by calculating the similarity between the latent gesture representation in the low-dimensional space and the real text representation in the abstract space. The calculation formula is as follows:
\begin{equation}
\begin{aligned}
    SA = cos(avg\_pool(V_{g}(G_{pred})),avg\_pool(V_{s}(S))),\\
\end{aligned}
\end{equation}
where \textit{G} represents the gestures generated by the model, and \textit{S} denotes the hidden states encoded by the BERT~\cite{devlin2018bert} model after tokenizing the transcribed text, serving as a representation of the semantics.

\noindent\textbf{Implementation Details.} 
Our system was trained on PyTorch with a denoiser learning rate of 3e-5 and a discriminator learning rate of 1.25e-4. The discriminator's gradient penalty term was set to 0.02, in line with DDGAN~\cite{xiao2021tackling}, and the CFG weight was set to 3.5. All models were trained on an A100 GPU for a uniform 1.3 million-step iteration, taking a total of 5 days. During evaluation, all methods were tested on a single V100 GPU for fairness. 
\subsection{Comparison with Contemporary Methods}
We present the quantitative results for speaker 2's test sequences in the BEATX dataset using the audio-to-gesture method in Table~\ref{tab:obj}. The purpose is to provide a fair comparison with the values reported in the original EMAGE paper. \textbf{\textit{\textcolor{black}{For a more comprehensive view of the quantitative experiments, we provide the quantitative results for the entire dataset in parentheses}}}. Our findings demonstrate that, in comparison with diffusion-based methods such as DSG~\cite{yang2023diffusestylegesture}, FreeTalker~\cite{yang2024freetalker}, and DiffGesture~\cite{zhu2023taming}, our approach outperforms them in terms of gesture matching, even with a 20-fold reduction in denoising steps. Moreover, our method surpasses VAE and VQ-VAE-based approaches like EMAGE~\cite{liu2023emage}, TalkShow~\cite{yi2023generating}, and CAMN~\cite{liu2022beat} in terms of beat alignment and diversity.

\begin{table}
  \resizebox{1.0\linewidth}{!}{
\begin{tabular}{lccccccccc}
\toprule[1.5pt]
\multirow{2}{*}{Method} & \multicolumn{5}{c}{BEATX} \\ \cline{2-6}
                        & FGD$\downarrow$  & SA$\uparrow$  & BA$\uparrow$  & DIV$\uparrow$      & steps \\  \hline
HA2G~\cite{liu2022audio}  &12.32  &0.13   &6.77 &8.626  &-\\
DisCo~\cite{liu2022disco} &9.417 &0.09   & 6.439 &9.912 &-\\
CaMN~\cite{liu2022beat}  &6.644 &0.22   &6.769 &10.86 &- \\
TalkShow~\cite{yi2023generating} &6.209 &0.22  &6.947 &13.47 &- \\
EMAGE~\cite{liu2023emage} &5.512(7.305) &0.17  &7.724(7.709) &10.88(10.948) &-\\\hline
DiffGesture(re-train)~\cite{zhu2023taming}  &12.8 &0.07 &7.08 &11.30  &1000  \\
DSG(re-train)~\cite{yang2023diffusestylegesture} &8.811(11.742) &0.08  &7.241(7.3368) &11.49(11.121) &1000\\
FreeTalker(re-train)~\cite{yang2024freetalker}    &7.712 &0.19 &7.73  &10.62 &1000 \\
\hline
\textbf{HoloGest(Ours)}&\textbf{5.3407}(6.457) &\textbf{0.66} &\textbf{7.957}(8.0281)&\textbf{14.15}(13.525) &50 \\
\hline
\end{tabular}
}
\caption{Objective metrics on BEATX. EMAGE provides the FGD evaluation model, where a lower value indicates a closer approximation to the original motion distribution. The calculation methods for BA and DIV are consistent with EMAGE. Steps represent the denoising steps in diffusion-based generation methods. \textbf{\textit{\textcolor{black}{The values in parentheses represent the evaluation results for the entire BEATX dataset}}}.}
\label{tab:obj}
\end{table}

However, gesture evaluation is subtle, and the FGD metric only reflects the similarity between generated results and the original distribution, not the actual effect of the virtual speaker or the trajectory and global rotation of holistic co-speech gestures. While EMAGE has metrics close to our system, it lacks prior knowledge of the entire motion sequence, leading to discord in its generated results, including unnatural global flips and severe skating phenomena. Its fingers also lack rich movements due to the absence of finger priors.

In contrast, our system, which introduces motion priors, generates reliable global movements without affecting vivid gestures, provides stable locomotion without unnatural flips or severe arm jittering, and offers users a more natural and harmonious experience. Table~\ref{tab:obj1} presents the physical metrics and semantic alignment scores, consistent with our observed phenomena.

\begin{table}
  \resizebox{1.0\linewidth}{!}{
\begin{tabular}{lcccc}
\toprule[1.5pt]
\multirow{2}{*}{Method} & \multicolumn{4}{c}{BEATX} \\ \cline{2-5}
                        & FGD$\downarrow$     & Skate$\downarrow$  & Float$\downarrow$ & SA$\uparrow$     \\  \hline
Real & 1.7e-4 & 0.0866 & 8.8015 & 0.82\\
EMAGE & 5.51 & 0.7904 & 34.6534 & 0.17\\
DSG & 8.811 & 0.4192 & 22.7526 & 0.08\\
\hline
\textbf{HoloGest(Ours)} & 5.34 & 0.1068 & 9.6317 & 0.66\\
\hline
\end{tabular}
}
\caption{Objective Metrics. Skate represents the skating metric when in contact with the ground, with values closer to Real being better. Float indicates the floating error during ground contact, with values closer to GT being better.}
  \label{tab:obj1}
\end{table}

\subsection{Qualitative Comparison} 
\begin{figure}
\center
  \includegraphics[width=1.0\linewidth]{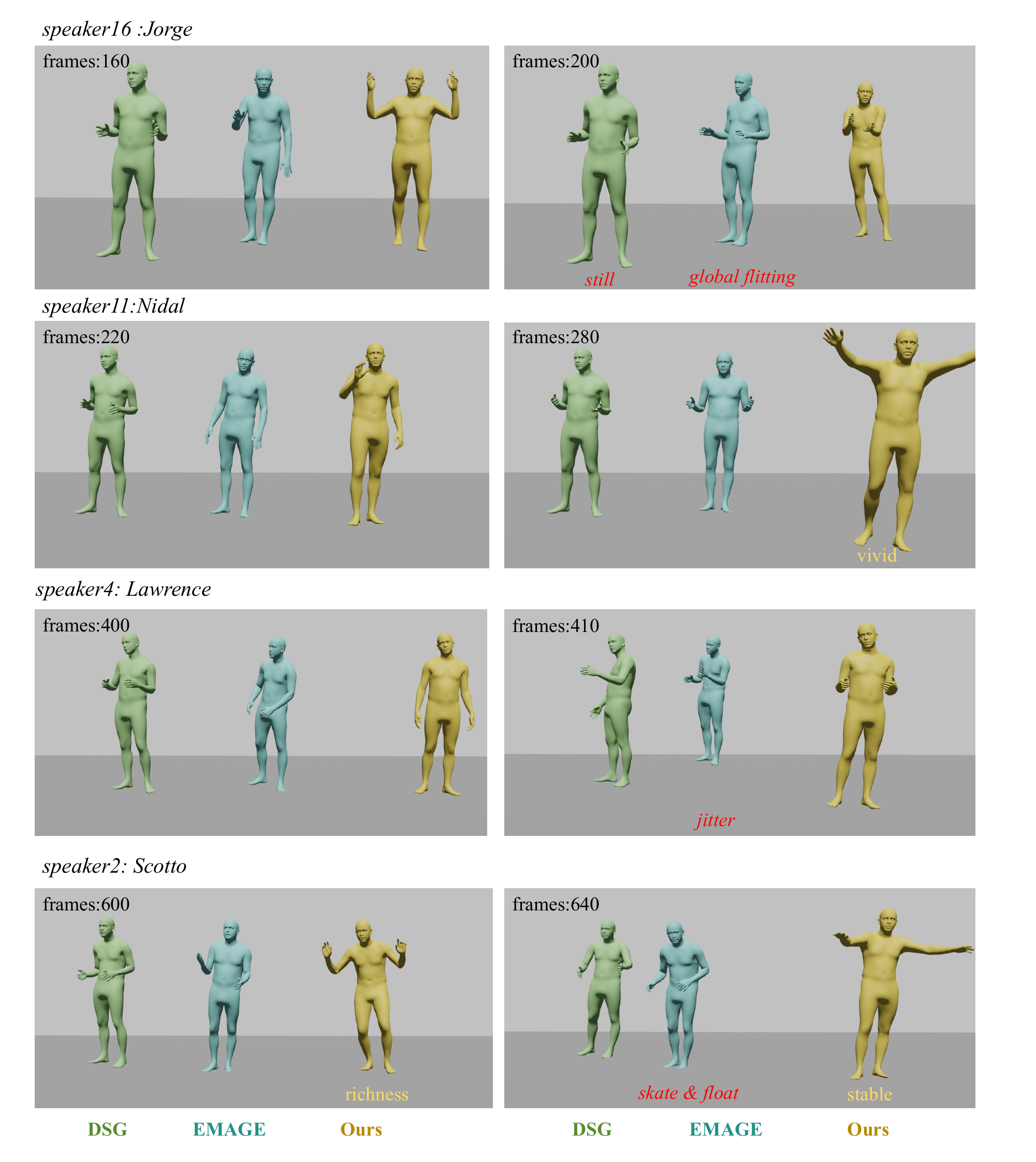}
  \caption{
  A comparison of three methods: DSG, a diffusion-based co-speech gesture generation method using DDPM (stiff limbs, slow inference, physically unnatural); EMAGE, an autoregressive generation method using VAE (motion artifacts, global flipping, physically unnatural); and our proposed generation method (rich movements, lively fingers, physically natural). We test on a sequence of an English-speaking presenter selected from BEATX. Red annotations indicate defects, while yellow annotations highlight advantages.
  }
  \label{fig:app}
\end{figure}

We present the results generated by DSG, EMAGE, and HoloGest on the BEATX test set. As seen in Figure~\ref{fig:app}, DSG's gesture generation lacks expressiveness, showing little movement during flat speech, resulting in a stiff appearance with unnatural phenomena like sliding and floating. The 1000-step DDPM sampling strategy also leads to inefficient generation. EMAGE, using VAE for direct regression, is fast but prone to motion artifacts and global flipping, affecting user experience.

In contrast, our model achieves large strides with fewer denoising steps, enhancing generation speed while maintaining high fidelity, making it suitable for real-time applications. The introduction of motion priors improves global motion and physical naturalness. Thanks to the semantic alignment module in the abstract space, our method is highly expressive, with large, lively upper limb movements and natural, rich lower limb movements. Our divide-and-conquer approach enhances the richness of finger movements and the stability of global actions.

\subsection{Ablation Study}
To validate the importance of each module in the system, we compare various variants obtained from the complete method:
\begin{itemize}
  \item \textit{\textbf{Baseline,}} directly decouples three body parts: the upper body, lower body, and fingers, as three parallel sub-models for denoising distribution modeling, while maintaining the DDPM denoising process, using only audio features as guiding signals, as done in DSG.
  \item \textit{\textbf{+ SIDD,}} to alleviate the inefficiency in generation caused by body decoupling, we introduce a semi-implicit denoising process that directly models the complex large-step denoising distribution, achieving acceleration by reducing the number of denoising steps.
  \item \textit{\textbf{+ SA,}} by employing the JEPA strategy, we use semantic features that are aligned with the real gesture sequences in the abstract space as additional guiding conditions, and independently conditionally tokenize them, similar to what is done in locomotion.
  \item \textit{\textbf{+ Global,}} in order to establish connections between the decoupled parts, we associate the features of the three parts and further map them to a single global token, serving as an additional global perceptual information.
  \item \textit{\textbf{+ Prior,}} incorporate global trajectory motion priors and finger priors as pre-trained models for secondary generation.
\end{itemize}

\begin{table}
  \resizebox{1.0\linewidth}{!}{
\begin{tabular}{lccccccc}
\toprule[1.5pt]
\multirow{2}{*}{Method} & \multicolumn{6}{c}{BEATX} \\ \cline{2-7}
                        & FGD$\downarrow$  & Skate$\downarrow$  & Float$\downarrow$ & SA$\uparrow$  & BA$\uparrow$  & DIV$\uparrow$ & steps \\  \hline
Real & 1.7e - 4 & 0.0866 & 8.8015 & 0.91 & - & - & - \\
DSG & 8.811 & 0.4192 & 22.7526 & 0.08 & 7.241 & 11.49 & 1000 \\
EMAGE & 5.512 & 0.7904 & 34.6534 & 0.17 & 7.724 & 10.88 & - \\
\hline
Baseline & 7.718 & 0.3922 & 19.7831 & 0.20 & 7.432 & 12.83 & 1000 \\
+ SIDD & 7.016 & 0.5567 & 25.1263 & 0.22 & 7.135 & 14.12 & 50 \\
+ SA & 6.351 & 0.5239 & 17.6612 & 0.60 & 7.946 & 14.26 & 50 \\
+ Global & 5.86 & 0.3396 & 19.023 & 0.66 & 7.953 & \textbf{14.29} & 50 \\
+ Prior & \textbf{5.3407} & \textbf{0.1068} & \textbf{9.6317} & 0.66 & \textbf{7.957} & 14.15 & 50 \\
\hline
\end{tabular}
}
\caption{Ablation study results on the module design in the system.}
\label{tab:abl}
\end{table}

Table~\ref{tab:abl} shows the ablation study results, with the complete system outperforming all ablation versions. The diffusion generative model with only decoupled structure shows some improvement compared to DSG but has a noticeable disadvantage in metrics compared to VAE-based methods. This is due to the lack of connections between parts, causing uncoordinated overall gestures when directly merged. Introducing global associations and semi-implicit denoising process alleviates this issue and improves generation efficiency. The introduction of semantic alignment features significantly enhances the richness of generated actions. Despite the improvements in metrics and gesture quality, the lack of global motion prior knowledge still leads to physically unnatural factors like skating and global jitter, impacting user experience. By introducing global trajectory priors and finger priors as pre-trained models for secondary generation, we achieve physically plausible results and provide users with a better experience. 

\subsection{User Study}
We used four human perceptual consistency scoring metrics as described in \cite{alexanderson2023listen}. These metrics evaluate human likeness (HL), speech-gesture appropriateness (SGA), gesture richness, and whole-body stability. To assess our method's visual performance, we conducted a user study on gesture sequences generated by each method. Evaluation segments varied from 16 to 40 seconds in length, averaging 26.2 seconds. We engaged 30 participants and used a scoring range of 1 to 5, with labels from "poor" to "excellent". Table 5 shows the average user opinion scores. We compared the results generated by the original DSG, EMAGE, our system without Prior, and the complete system. As per user feedback~\ref{tab:user}, our method generates high-quality co-speech gesture sequences comparable to, or better than, real data, and does so faster than traditional DDPM diffusion generative methods. Notably, our method takes only 0.88 seconds to generate a 2-second gesture sequence, compared to approximately 7 seconds using 1000-step DDPM, making it suitable for real-time applications like human-computer communication.

\begin{table}
  \resizebox{1.0\linewidth}{!}{
    \begin{tabular}{lcccc}
      \toprule[1.5pt]
      \multirow{2}{*}{Method} & \multicolumn{4}{c}{BEATX} \\ \cline{2-5}
                              & HL$\uparrow$ & SGA$\uparrow$ & R$\uparrow$ & Stable$\uparrow$ \\ \hline
      Real & $4.61 \pm 0.17$ & $4.72 \pm 0.20$ & $4.66 \pm 0.07$ & $4.89 \pm 0.02$ \\
      DSG & $3.70 \pm 0.12$ & $3.91 \pm 0.14$ & $4.27 \pm 0.15$ & $3.12 \pm 0.12$ \\
      EMAGE & $3.44 \pm 0.18$ & $4.11 \pm 0.14$ & $3.56 \pm 0.09$ & $2.87 \pm 0.22$ \\ \hline
      \textbf{HoloGest(Ours)} & $4.47 \pm 0.09$ & $4.51 \pm 0.19$ & $4.82 \pm 0.1$ & $4.71 \pm 0.11$ \\ \hline
    \end{tabular}
  }
  \caption{95\% Confidence Interval for User Study Average Score.}
  \label{tab:user}
\end{table}

\section{Conclusions}
In this study, we tackled challenges in generating holistic co-speech gestures. By innovating upon diffusion-based methods with implicit marginal constraints and explicit auxiliary forward diffusion regularization, our model enabled faster inference and mitigated generation speed inefficiencies. Additionally, we considered motion prior and introduced a pre-trained model on extensive human motion data, generating physically accurate gesture sequences and enhancing user experience. Our approach significantly accelerated HoloGest's generation while maintaining high fidelity, paving the way for future real-time synchronous gesture generation tasks.

{
    \small
    \bibliographystyle{ieeenat_fullname}
    \bibliography{main}
}
\end{document}